# Multi-Objective Optimization of the Textile Manufacturing Process Using Deep-Q-Network Based Multi-Agent Reinforcement Learning

ZHENGLEI HE    KIM PHUC TRAN    SEBASTIEN THOMASSEY    XIANYI ZENG

*ENSAIT, GEMTEX – Laboratoire de Génie et Matériaux Textiles, F-59000 Lille, France*

JIE XU, CHANGHAI YI

*Wuhan Textile University, 1$^{st}$, Av Yangguang, 430200, Wuhan, China*

*National Local Joint Engineering Laboratory for Advanced Textile Processing and Clean Production, 430200, Wuhan, China*

*Abstract:* Multi-objective optimization of the textile manufacturing process is an increasing challenge because of the growing complexity involved in the development of the textile industry. The use of intelligent techniques has been often discussed in this domain, although a significant improvement from certain successful applications has been reported, the traditional methods failed to work with high-dimension decision space and require prior experts' knowledge as well as human intervention. Upon which, this paper proposed a multi-agent reinforcement learning (MARL) framework to transform the optimization process into a stochastic game and introduced the deep Q-networks algorithm to train the multiple agents. A utilitarian selection mechanism was employed in the stochastic game, which maximizes the sum of all agents' rewards (obeying the increasing ε-greedy policy) in each state to avoid the interruption of multiple equilibria and achieve the correlated equilibrium optimal solutions of the optimizing process. The case study result reflects that the proposed MARL system is possible to achieve the optimal solutions for the textile ozonation process and it performs better than the traditional approaches.

*Keywords:* Deep Reinforcement Learning; Deep Q-Networks; Multi-objective; optimization; Decision; Process; Textile Manufacturing.

1. Introduction

The textile manufacturing process adds value to fiber materials by converting the fibers into yarns, fabrics, and finished products. But it is very difficult to decide in the scenario of a textile manufacturing process owing to the complicated application of a wide range of dependent and independent parameter variables in a very long chain of the textile processes [1]. Meanwhile, the optimization of a textile manufacturing process in terms of scenario solution always involves the consideration of multiple objectives as the performance of a textile manufacturing process generally is governed by a few conflicting criteria with different significance [2].

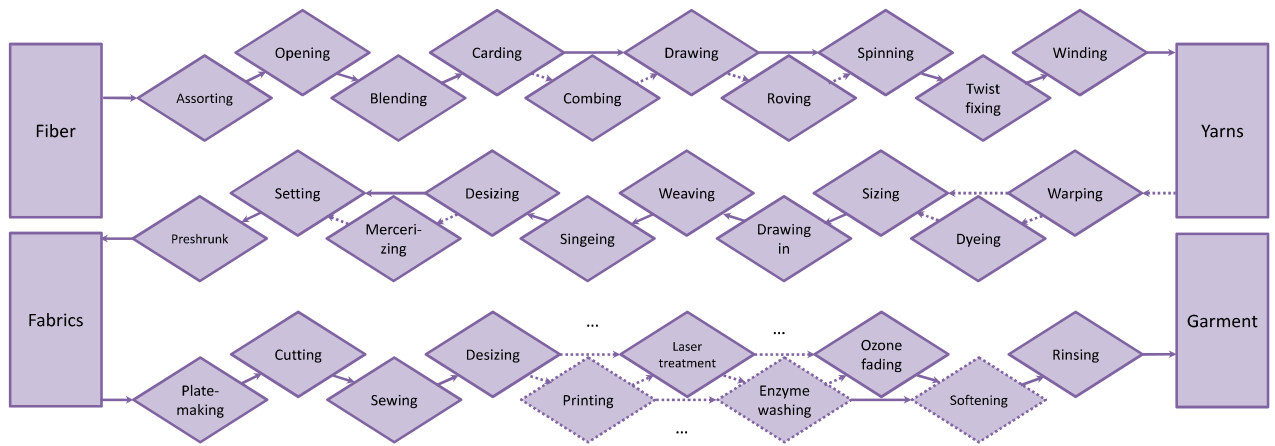

Figure 1. A general illustration of the textile manufacturing processes from fiber to garment

In recent years, multi-objective optimization of the textile manufacturing process has drawn increasing attention thanks to the rapid development and wide application of intelligent techniques in the textile industry[3–6]. For example, Sette and Langenhove [7] simulated and optimized the fiber-to-yarn process to balance the conflicting targets of cost and yarn quality. Majumdar et al. [8] optimized the functional clothing in terms of ultraviolet protection factors and air permeability.

However, since textile manufacturing consists of a variety of processes (a general illustration of the textile manufacturing processes is displayed in Figure 1), the probable combinations of processes and parameters could be stochastic and enormous when the factors of the performance vary in many respects. Meanwhile, the known and unknown factors cannot be interpolated and extrapolated reasonably based on experimental observations or mill measurements due to the shortage of knowledge on the evaluation of the interaction and significance at weight contributing from each variable [9]. It turns out that classical techniques are no longer efficient in some scenarios of textile manufacturing on account of the growing complexity. By contrast, intelligent techniques, especially the machine learning algorithms, have been successfully applied in many sectors and shown their power for coping with complicated optimization problems with large-scale data and high-dimension searching space [10], which is strongly related to the nature of the present problem of interest.

Historically, researchers generally dealt with the multi-objective optimization problems of textile manufacturing using mathematical programming methods [8,11] and meta-heuristic algorithms [12]. These approaches, however, either simplify the case by omitting certain non-essential details to achieve manageable equations based on scarification on the accuracy or require prior experts' knowledge and human intervention. More importantly, they fail to work with high-dimension decision space. It is noticed that the novel machine learning algorithms are demonstrating increasingly versatility and power in the practical applications of optimization issues in the industry, and considerable researches paid attention to using reinforcement learning (RL) algorithms[13–19] in this regard. However, most of these previous RL studies focused only on single-objective problems.

It is known that multi-objective optimization problem could be transformed into game-theoretic models to be well solved [20,21], and recent developments of a multi-agent system for optimizing multiple objectives on the basis of game theory have shown its extreme capability of dealing with functions having high dimensional space [22,23]. On the other hand, the multi-agent reinforcement learning (MARL) has been proposed by many contributions for robotics distributed control, telecommunications, traffic light control, and dispatch optimization, etc. [24–26], but traditional MARL algorithms generally can hardly handle the large-scale problem, the applicability of it was therefore very limited [27]. While in recent years, the development of deep reinforcement learning (DRL) has achieved many outstanding results, which prompts a growing number of research efforts paying to the investigations of algorithms and applications of DRL in MARL environment [28–30]. Although studies reported the use of MARL and DRL for optimizing workflow scheduling, electronic auctions and traffic control problems with multiple objectives [31], very limited work solved a complex production problem, especially in the textile manufacturing industry.

Upon which, this paper formulates the multi-objective optimization problems of the textile manufacturing process into a Markov game paradigm, and collaboratively applying multi-agent deep-Q-networks (DQN) reinforcement learning instead of current methods to address it.

## 2. State of art and contributions

Textile manufacturing originates from the fibers (e.g. cotton) to final products (such as curtain, garment, and composite) through a very long procedure with a wide range of different processes filled with a large number of variables. The simultaneous optimization of multiple targets in a textile production scheme from the high dimensional space is extremely challenging with a dramatically high cost.

There have been a variety of works on the textile process multi-objective optimization from the last decades. For example, Sette and Langenhove [7] simulated and optimized the fiber-to-yarn process to balance the conflicting targets of cost and yarn quality. Majumdar et al. [8] optimized the functional clothing in terms of ultraviolet protection factor and air permeability. Mukhopadhyay et al. [32] attempted to optimize the parametric combination of injected slub yarn to achieve the least abrasive damage on fabrics produced from it. Almetwally [33] optimized the weaving process performances of tensile strength, breaking extension, and air permeability of the cotton woven fabrics by searching optimal parameters of weft yarn count, weave structure, weft yarn density and twist factor. These works generally used the prior techniques that combine the multiple objectives into a single weighted cost function, the classical approaches such as weighted sum, goal programming, min-max, etc. are not efficient due to the fact that they cannot find the multiple solutions in a single run but times as many as the number of desired Pareto-optimal solutions. Pareto optimal solutions or non-dominated solutions are equally important in the search space that superior to all the other solutions when multiple objectives are considered simultaneously, and the curve formed by joining Pareto optimal solutions is the well-known Pareto optimal front [34].

Heuristic and meta-heuristic algorithms are a branch of optimization in computer science and applied mathematics[35]. The investigations and applications of the related algorithms and computational complexity theory are very popular in the textile manufacturing industry with regard to the multi-objective optimization that is feasible to approach the Pareto optimal solutions. Among these, evolutionary algorithms such as genetic algorithms (GA) and gene expression programming (GEP) are the ones that are most often taken into consideration in previous studies in the textile sector. Kordoghli et al. [36] schedule the flow-shop of a fabric chemical finishing process aiming at minimal make-span and arresting time of machine simultaneously using multi-objective GA. Nurwaha [37] optimized the electrospinning process performance in terms of fiber diameter and its distribution by searching for optimal solutions with regard to the processing parameters including solution concentration, applied voltage, spinning distance and volume flow rate. The electrospinning process parameters were mapped to the performances by the GEP model, and a multi-objective optimization method was proposed on the basis of GA to find the optimal average fiber diameter and its distribution. Wu and Chang [38] proposed a nonlinear integer programming framework on the basis of GA to globally optimized the textile dyeing manufacturing process. The results of their case study presented the applicability and suitability of this methodology in a textile dyeing firm and exactly reflected the complexity and uncertainty of application challenges in the optimal production planning program in the textile industry.

In terms of multi-objective optimization, the general GA systems developed in the works above may not efficient in certain cases as the elitist individuals could be over-reproduced in many generations and lead to early convergence. To this end, Deb [39] proposed a Non-dominated sorting genetic algorithm Ⅱ (NSGA-Ⅱ) that introduced a specialized fitness function and fast non-domination sorting as well as crowding distance sorting in the common GA system to promote solution diversity in the generations. Such a modified strategy has been widely applied in related textile studies. For instance, Ghosh et al. [40] optimized the yarn strength and the raw material cost of the cotton spinning process simultaneously with NSGA-Ⅱ on the basis of two objective function models in terms of artificial neural networks and regression equation. Similarly, Muralidharan et al. [41]described the combined use of NSGA-Ⅱ with response surface methodology for the design and control of color fastness finish process to optimize five quality characteristics, i.e. shade variation to the standard, color fastness to washing, center to selvedge variation, color fastness to light and fabric residual shrinkage. Majumdar et al. [42] derived the Pareto optimal solutions using NSGA-II to obtain the effective knitting and yarn parameters to engineer knitted fabrics having optimal comfort properties and desired level of ultraviolet protection. Barzoki et al. [43] and Vadood et al. [44] employed this algorithm with artificial neural networks and Fuzzy logic respectively to optimize the properties of core-spun yarns in the rotor compact spinning process, where the investigated process parameters consist of the filament pre-tension, yarn count and type of sheath fibers, and the objectives were yarn tenacity, hairiness and abrasion resistance for the former but elongation and hairiness for the latter respectively. Apart from the GA frameworks, applications reported of other heuristic or meta-heuristic algorithms for multi-objective optimization in the textile domain also have been presented with synergetic immune clonal selection (SICS), artificial bee colony (ABC) algorithm, ant colony optimization (ACO), and particle swarm optimization (PSO) [12,45]. Meanwhile, simultaneous optimization using the desirability function [46], in addition to the heuristic or meta-heuristic algorithms, was very popular in the textile manufacturing process multi-objective optimization applications as well [47,48].

As the focus of this paper is on the large-scale data and high-dimension space in the multi-objective optimization problems in the textile manufacturing process, however, the aforementioned techniques are significantly restricted by prior expert's knowledge from a global point of view and failed to work with large-scale data and high-dimension decision space. Although these classical techniques might make sense in certain previous studies, the effectiveness and efficiency of these traditional tools would be unacceptable in the industry 4.0 era with the massive quantities of data as well as the high complexity of the textile manufacturing process. For example, the heuristic methods are time-consuming that can hardly be applied in the context of industrial practice, when the number of variables is very large, along with large change intervals [49]. By contrast, multi-agent reinforcement learning (MARL) is a machine learning approach using a relatively well understood and mathematically grounded framework of Markov decision process (MDP) based on game theory that has been broadly applied to tackle the practical multi-objective optimization problems in the industry [26,31]. However, at present, none of the studies ever reported the use of MARL for optimizing the textile manufacturing process before. Unlike classical RL algorithms such as Q-learning and State–action–reward–state–action (SARSA) that relied on a memory-intensive tabular representation of the value or instant reward, which not only perform inefficiently in the high-dimensional cases but waste computational power as well, deep-Q-networks (DQN) algorithm utilizes deep learning tools and strategies of experience replay [50] and fixed Q-target that is quite good at coping with the large-scale issues and has recently been well evaluated in many applications of deep reinforcement learning (DRL) [18,51,52]. Given the advantages that the DQN can offer when confronted with high-dimensional complexities in the textile process, this paper proposes the DQN algorithm in the MARL framework for multi-objective optimization. To the best of our knowledge, this is the first paper that investigates the multi-objective textile process optimization problems using DQN based MARL system.

The main contributions of this paper are listed below:

(1) Construction of a machine learning-based multi-objective optimization system for the textile manufacturing process.
(2) Formulation of optimizing the textile manufacturing process as a Markov decision process, and applying reinforcement learning to solve the problem.
(3) Transforming the multi-objective optimization problems of textile manufacturing into the game-theoretic model, and introducing multi-agent for searching the optimal process solutions.
(4) The application of DQN is extended to the multi-agent reinforcement learning system. Compared to the tabular RL algorithms applied in prior related works, DQN is more applicable and preferred to cope with the complicated realistic problem in the textile industry.

The rest of this paper is organized as follows: Section 3 presents the problem formulation of textile manufacturing process multi-objective optimization and the mathematical representation of the problem in the system model, followed by the detailed illustrated framework of the proposed MARL system and a case study of applying the system to optimize an advanced textile finishing process optimization in Section 4 and Section 5 respectively. Finally, conclusions and future works are discussed in Section 6.

3. **System model**

Consider the solution of a textile manufacturing process $P$ is composed and determined by a set of parameter variables $\{v_1, v_2... v_n\}$, the impacts of these variables on the process performance could be varied a lot from $n$ different respects with uncertainty, as the number of the processes and the related variables in the textile manufacturing industry is enormous and the influences of these variables on the targeted optimization performance are unclear. For example, the longer time was taken of a textile process generally would lead to the increment of production cost, and a tiny enhance of temperature used in the textile production process could significantly arouse the power consumption, but sometimes the enhanced temperature may promote the process efficiency so that decrease the production cost eventually. Therefore, it is necessary to study the interrelated effects of process variables on process performance. From the engineering perspective, it is important to achieve a solution in the textile manufacturing process that can achieve good quality and avoid idle time, waste and pollutions at the same time. Models that incorporate the information of the process simulating the variation of multiple objective performances from the change of variable in the solutions are rather essential.

Suppose models exist that can map variables $v_1, v_2... v_n$ of the process solution $P$ to its performance in accordance with $m$ objectives, the performance of a specific solution could be simulated by:

$$f_i(P) = f_i(v_1, v_2 ... v_n) \quad for \quad i = 1, ... m \tag{1}$$

When a decision-maker who wants to find a solution that satisfies $m$ objectives of the process performances that the objectives are non-commensurable and no preference of the objectives related to each other is coming up with the decision-maker. The multi-objective problem could be defined as giving the $n$-dimensional variable vector $P = \{v_1, v_2... v_n\}$ in the solution space, finding a vector of $p^*$ that optimizes a given set of $m$ objective functions:

$$f(p^*) = \{f_1(p^*), f_2(p^*), ..., f_m(p^*)\} \tag{2}$$

The solution space is generally restricted by a series of constraints, when the domain of $v_j \in V_j$ for $j = 1, ..., n$ is known, and representing the $m$ objectives by $M$, the objective of the problem is to find (3):

$$argmax_{v_j \in V_j}[\ f(v_1, v_2 \ldots v_n)\ |\ M] \qquad \text{for } j = 1, \ldots, n \tag{3}$$

Equation (3) aims at searching the optimal solution of variable settings, while there are always conflicting objectives that satisfying one single target but lead to unacceptable results to the others. A perfect multi-objective solution that simultaneously optimizes each objective function is almost impossible. To this end, this paper proposes a self-adaptive DQN-based MARL framework where the *m* optimization objectives are formulated as *m* DQN agents that trained through a self-adaptive process constructed upon a Markov game.

**4. Methodology**

**4.1.** *Multi-objective optimization of textile manufacturing process as Markov game*

We begin by formulating the single objective textile process optimization problem as a Markov decision process (MDP) in terms of a tuple :{ *S, A, T, R*}, where *S* is a set of environment states, *A* is a set of actions, *T* is the state transition probability function, *R* is a set of reward or losses. An agent in an MDP environment would learn how to take action from *A* by observing the environment with states from *S*, according to corresponding transition probability *T* and reward *R* achieved from the interaction. The Markov property indicates that the state transitions are only dependent on the current state and current action is taken, but independent of all prior states and actions [53]. While in the case of a multi-agent system, the joint actions are the result of multiple agents, the MDP is generalized to the stochastic Markov game of {*S, $A^1, \ldots, A^m$, T, $R^1, \ldots, R^m$* }, where *S* and *T* are similar to the MDP that are the finite set of environment states and the state transition probability function respectively in a Markov game, whereas differently, *m* is the number of agents, $A^i$ for *i* =1,…, *m* are the finite sets of actions available to the agent *i*, $R^i$ for *i* =1,…, *m* are the reward functions of the agent *i*.

As known that the solution of a textile manufacturing process is affected by a number of variables as *P* {$v_1, v_2 \ldots v_n$}, if the possible value of $v_j$ is $h(v_j)$, the feasible values of the parameter in the process can define the environment space *S* from $\prod_{j=1}^{n} h(v_j), v_j \in V_j$ impacting the performance of the textile process with regard to the *k* objectives. These parameter variables are independent of each other and obey a Markov process that models the stochastic transitions from a state $S_t$ at time step *t* to the next state $S_{t+1}$, where the environment state at time step *t* is:

$$S_t = [\ s_t^{v_1}, s_t^{v_2} \ldots s_t^{v_n}] \in S \tag{4}$$

RL algorithm trains an agent to act optimally in a given multi-agent environment based on the observation of states and other agents as well as the feedback derived from the interactions, acquiring rewards and maximizing the accumulative future rewards over time from the interaction[53]. In our case, the agents learn in the interaction with the environment and other agents by taking action that can be conducted on the parameter variables $\in P$ {$v_1$, $v_2$... $v_n$} at time step $t$. Specifically, the action of an agent in a time step $t$ of optimizing a textile manufacturing process in the Markov game, could be adjusting variable $v_j$ to keep (0) or change to up (+) and down (-) with a specific unit $u_j$ subjected to the constraint. As a result, there are $3^n$ actions in total in the joint action space $A$ and, for simplicity, the action vector $A_t$ at time step $t$ could be:

$$A_t = [a_t^{v_1}, a_t^{v_2} ... a_t^{v_n}], \quad \text{where } a_t^{v_j} \in \{-u_j, 0, +u_j\}, v_j \in V_j \text{ for } j = 1, ..., n \tag{5}$$

We define $A = \prod_{i \in m, s \in S} A^i(s)$ for the joint action from overall the agent $i$'s set of actions at state $s$. The $m$ objectives of textile manufacturing process optimization are assigned to $m$ agents in this Markov game. As known that apart from the benefits derived from the distributed nature of the multi-agent system such as parallel computation, the experience sharing from different agents also significantly improve the multi-agent algorithms. Therefore, it is assumed that agents can observe each other's action and rewards to select the joint distribution in our case, and the joint action is determined by the actions selected of each agent $(A^1, ..., A^i, ..., A^m)$.

The state transition probabilities, as mentions that, are only dependent on the current state $S_t$ and action $A_t$. It specifies how the reinforcement agents take action $A_t$ at time step $t$ to transit from $S_t$ to next state $S_{t+1}$ in terms of $T$ $(S_{t+1} | S_t, A_t)$. For all $a_t^{v_j} \in \{-u_j, 0, +u_j\}, v_j \in V_j$, $T(S_{t+1} | S_t, A_t) > 0$ and $\sum_{S_{t+1} \in S} T(S_{t+1} | S_t, A_t) = 1$. The reward achieved by an agent in an environment is specifically related to its transition between states, which evaluates how good the transition agent conducts and facilitates the agent to converging faster to an optimal solution.

When the reinforcement agents perform a joint action $A_t$ at time step $t$ to divert the system from $S_t$ to next state $S_{t+1}$ with transition probability $T$, each agent would earn reward $R_i(S_t, A_t)$ from (3) of the objective functions. This procedure would be repeated at time $t+1$ again, and finally, converge agents' behaviors to a stationary policy. According to a previous study [54], a random forest (RF) predictive model is applied to simulate the textile process in this proposed framework and implement the objective functions (3) to earn the agents' rewards. As illustrated in Figure 2 the textile manufacturing process multi-objective optimization problem in the paradigm of MARL, the optimization objectives are abstracted as RL agents, given feedbacks from the RF models integrated with the Markov game environment with state-space formulated in Equation (4) that consist of all the parameter variables of the simulated textile process, the agents, are able to evaluate the values of its actions for adjusting the parameter variables with regard to the state (solution) and consequently improve its policy in the environment to optimize objectively gradually.

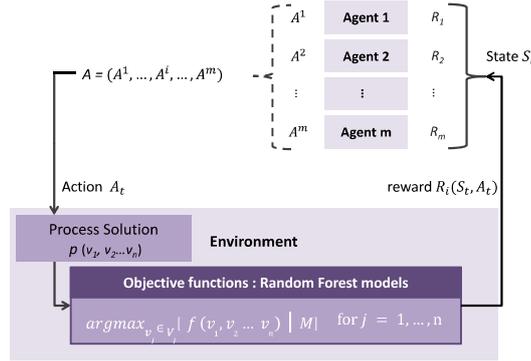

Figure 2. The Markov game for textile manufacturing process multi-objective optimization in the proposed framework

Stochastic games are neither fully cooperative nor fully competitive [24]. The performance of multi-objective optimization of our case in stochastic Markov game is determined by the agents' capability of gathering information about the other agents' behavior and the reward functions from the interaction to make a more informed decision thereafter. The rewards mechanisms along with the interaction among agents perform a significant function in this respect, so that the proposed system, similar to the study of [31], employs a utilitarian selection mechanism $h = argmax_{A \in \Delta(A(S))} \sum_{i \in M} Q_i(s,a)$ that maximize the sum of all agents' rewards in each state to avoid the interruption of multiple equilibria. Convergence to equilibria is a basic stability requirement of MARL, and the Nash equilibrium is a well-known solution concept for the stochastic game that a joint strategy leading to a status of no agent is incentive to change its strategy. But a correlated equilibrium with increased generality instead of Nash equilibrium is taken into consideration in this issue as it allows agents' strategies to be interdependent. It is a joint distribution of actions from which none of the agents has any motivation to deviate unilaterally. Consequently, the solutions of the textile manufacturing process multi-objective optimization problem are correlated equilibria.

Formally, given a Markov game, a joint stationary policy $\pi$ leads to a correlated equilibrium when:

$$\forall i \in M, s \in S \mid \sum_{a \in A^{-i}(s)} \pi_s \, Q_i^\pi(s, a) \geq \sum_{a \in A^{-i}(s)} \pi_s \, Q_i^\pi(s, a') \tag{6}$$

where $A^{-i}(s)$ is the set of action vectors in state *s* excluding ones of agent *i*. The above inequality denotes that in state *s*, when it is recommended that agent *i* play *a*, it prefers to play *a*, because the expected utility of *a* is greater than or equal to the expected utility of $a'$, for all $a'$.

### 4.2. *Deep Q-networks reinforcement learning algorithm*

Classical RL algorithms such as the Q-learning and the SARSA (0/λ), are based on a memory-intensive tabular representation (i.e. Q-table) of the value or the instant reward, of taking an action *a* in a specific state *s* (the Q value of state-action pair, a.k.a Q(s, a)). These tabular algorithms impede the RL in realistic large-scale applications due to the huge amounts of states or actions involved. The tabular expression not only comes short of recording all of the Q(s, a) in these applications but also shows poor generalization in the environment with uncertainty.

The deep neural networks (DNNs) is another widely applied machine learning technique coping with large-scale issues and has recently been innovatively combined with the RL to evolve toward deep reinforcement learning (DRL) algorithms. Deep-Q-network (DQN) is a DRL algorithm developed by Mnih et al. [52] in 2015 as the first artificial agent that is capable of learning policies directly from high-dimensional sensory inputs and agent-environment interactions. It is an RL algorithm proposed based on Q-learning, one of the most widely used model-free off-policy and value-based RL algorithms.

The Q-learning agent learns through estimating the sum of rewards *r* for each state $S_t$ when a particular policy $\pi$ is being performed. It uses a tabular representation of the $Q^\pi(S_t, A_t)$ value to assign the discounted future reward *r* of state-action pair at time step *t* in Q-table. The target of the agent is to maximize accumulated future rewards to reinforce good behavior and optimize the results. In the Q-learning algorithm, the maximum achievable $Q^\pi(S_t, A_t)$ obeys Bellman equation on the basis of an intuition: if the optimal value $Q^\pi(S_{t+1}, A_{t+1})$ of all feasible actions $A_{t+1}$ on state $S_{t+1}$ at the next time step is known, then the optimal strategy is to select the action $A_{t+1}$ maximizing the expected value of $r + \gamma \cdot max_{A_{t+1}} Q^\pi(S_{t+1}, A_{t+1})$.

$$Q^\pi(S_t, A_t) = r + \gamma \cdot max_{A_{t+1}} Q^\pi(S_{t+1}, A_{t+1}) \tag{7}$$

According to the Bellman equation, the Q-value of the corresponding cell in Q-table is updated iteratively by:

$$Q^\pi(S_t, A_t) \leftarrow Q^\pi(S_t, A_t) + \alpha[r + \gamma \cdot max_{A_{t+1}} Q^\pi(S_{t+1}, A_{t+1}) - Q^\pi(S_t, A_t)] \qquad (8)$$

where $S_t$ and $A_t$ are the current state and action respectively, while $S_{t+1}$ is the state achieved when executing $A_{t+1}$ in the set of *S* and *A* in any given MDP tuples of {*S, A, T, R*}. $\alpha \in [0, 1]$ is the learning rate, which indicates how much the agent learned from new decision-making experience ($Q^\pi(S_{t+1}, A_{t+1})$) would override the old memory ($Q^\pi(S_t, A_t)$). *r* is the immediate reward, $\gamma \in [0, 1]$ is the discount factor determining the agent's horizon.

The agent takes action on a state in the environment and the environment interactively transmits the agent to a new state with a reward signal feedback. The basic principle of Q-learning algorithm essentially relies on a trial and error process, but different from humans and other animals who tackle the real-world complexity with a harmonious combination of RF and hierarchical sensory processing systems, the tabular representation of Q-learning is not efficient at presenting an environment from high-dimensional inputs to generalize past experience to new situations [52].

Q-table saves the Q value of every state coupled with all its feasible actions in a given environment, while the growing complexity in the problem nowadays indicates that the states and actions in an RL environment could be innumerable (such as Go game). In this regard, DQN applies DNNs instead of Q-table to approximate the optimal action-value function. The DNNs feed by the state for approximating the Q-value vector of all potential actions, for example, are trained and updated by the difference between Q-value derived from previous experience and the discounted reward obtained from the current state. While more importantly, to solve the instability of RL representing the Q value using nonlinear function approximator [55], DQN innovatively proposed two ideas termed experience replay [50] and fixed Q-target. As known that Q-learning is an off-policy RL, it can learn from the current as well as prior states. Experience replay of DQN is a biologically inspired mechanism that learns from randomly taken historical data for updating in each time step, which therefore would remove correlation in the observation sequence and smooth over changes in the data distribution. Fixed Q-target performs a similar function, but differently, it reduces the correlations between the Q-value and the target by using an iterative update that adjusts the Q-value towards target values periodically.

Specifically, the DNNs approximate Q-value function in terms of *Q-*(*s, a; $\theta_i$*) with parameters $\theta_i$ which denotes weights of Q-networks at iteration *i*. The implementation of experience replay is to store the agent's experiences $e_t=$ (*$S_t$, $A_t$, $r_t$, S $_{t+1}$,*) at each time step *t* in a dataset $D_t$ = {$e_1$,…$e_t$,}. Q-learning updates were used during learning to samples of experience, (S, A, r, S') ~ U(D), drawn uniformly at random from the pool of stored samples. The loss function of Q-networks update at iteration *i* is:

$$L_i(\theta_i) = \mathbb{E}_{(S,A,r,S') \sim U(D)}\left[\left(\left(r + \gamma \cdot \max_{A'} Q(S', A'; \theta_i^-)\right) - Q(S, A; \theta_i)\right)^2\right] \quad (9)$$

where $\theta_i^-$ are the network weights from some previous iteration. The targets here are dependent on the network weights; they are fixed before learning begins. More precisely, the parameters $\theta_i^-$ from the previous iteration is fixed as optimizing the $i_{th}$ loss function $L_i(\theta_i)$ at each stage and are only updated with $\theta_i$ every *F* steps. To implement this mechanism, DQN uses two structurally identical but parametrically differential networks, one of it predicts $Q(S, A; \theta_i)$ using the new parameters $\theta_i$, the rest one predicts $r + \gamma \cdot \max_{A'} Q(S', A'; \theta_i^-)$ using previous parameters $\theta_i^-$. Every *F* steps, the *Q* network would be cloned to obtain a target network $\hat{Q}$, and then $\hat{Q}$ would be used to generate Q-learning target $r + \gamma \cdot \max_{A'} Q(S', A'; \theta_i^-)$ for the following *F* updates to network *Q*.

### 4.3. *DQN based MARL for multi-objective optimization of textile manufacturing process*

The pseudo-code of the DQN based MARL framework for multi-objective optimization of the textile manufacturing process is illustrated in Algorithm 1. Correspondingly, Figure 3 graphically depicts a single episodic running of Algorithm 1. To learn a correlated equilibrium strategy, the DQN agents interact with the textile solution environment and other agents iteratively on the basis of local updates of Q-values and policy at each state. As mentioned, the random forest models (RF) are constructed to simulate the objective performances of the textile process in the proposed framework. Along with suitable reward mechanisms designed according to objective functions (in our framework, the reward of an agent is given by the improvement of the objective performance from the current state compared with the last state), the convergence of the DQN-based algorithm in multi-agent settings can be guaranteed.

The given algorithm can work without episodes as the target of agents is to find the optimized solution, in terms of state in the environment with specifications of multiple objective performances in the textile process simulated by RF models, however, the lack of exploration of the agent in an environment may cause local optimum in a single running. So we initialize the first state randomly from each sub-state $s_t^{v_i}$, where parameter variables $v_j \in V_j$, and introduce an episodic learning process to the agent for enlarging the exploration and preventing local optimum.

Additionally, we employed an increasing $\varepsilon$-greedy policy to balance the exploration and exploitation of states at the learning period and optimizing period respectively. As illustrated in Algorithm 2, increasing $\varepsilon$-greedy is employed with an increment given in each time step from 0 until it equals to $\varepsilon_{max}$. This helps the agents find the best actions in the present state to go to the next state with a possibility of $\varepsilon$ that may also randomly choose an action with a possibility of $1 - \varepsilon$ to get a random next state. In this regard, the agents can explore the unexplored states without staying in the exploitation of already experienced states of Q-networks, and plentifully exploit them when the states are traversed enough.

**Algorithm 1: DQN based MARL main body:**

**Input:** game $\Gamma$, RF models for simulating $m$ objective performance $f(f_1 ... f_m)$, selection mechanism $h$, expected performance of process $P(p_1, p_2 ... p_m)$, number of episodes $E$, number of time steps $N$, learning rate $\alpha$, discount factor $\gamma$, the step updating DQN $F$, replay memory size $D$;

Initialize function $Q$ with random weights $\theta$;

Initialize function $\hat{Q}$ with weights $\theta^- = \theta$;

Initialize state $s_0 = (v_1, v_2 ... v_n)$

**For** episode =1, $E$ **do**

  **For** time step=1, $N$ **do**

    Choose an action randomly or $a_t \in h$ using increasing $\varepsilon$-greedy policy

    Execute action $a_t$, observe next state $s_{t+1}$

    Estimate $f_1(s_t) ... f_m(s_t)$ and $f_1(s_{t+1}) ... f_m(s_{t+1})$ to observe $r_t$ ($r_t = \sqrt{(f_i(s_t) - p_i)^2} - \sqrt{(f_i(s_{t+1}) - p_i)^2}$)

    Store transition ($s_t, a_t, r_t, s_{t+1}$) in $D$

    Sample random minibatch of transitions ($s_t, a_t, r_t, s_{t+1}$) from $D$

    Set $y_i = \begin{cases} r_j & \text{if terminates at step } j + 1 \\ r_j + \gamma max_{a'} \hat{Q}(s_{j+1}, a'; \theta^-) & \text{otherwise} \end{cases}$

    Perform a gradient descent step on $\left(y_i - Q(s_j, a_j; \theta)\right)^2$ with regard to $\theta$

    Every $R$ steps reset $\hat{Q} = Q$

    $s_t \leftarrow s_{t+1}$

**End For**

**End For**

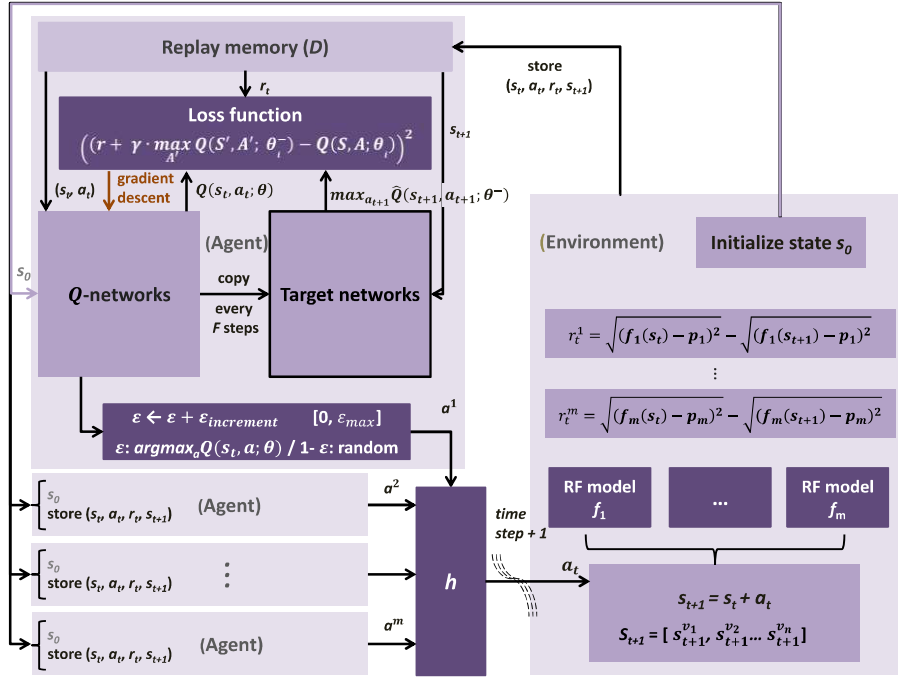

Figure 3. Flowchart of the algorithm implementing the proposed DQN based multi-agent system for optimizing textile manufacturing process with multiple objectives

---

**Algorithm 2:** Increasing $\varepsilon$-greedy policy

---

**Input:** $\varepsilon_{increment}$, $\varepsilon_{max}$

$\varepsilon \leftarrow \varepsilon + \varepsilon_{increment}$ $(0 \leq \varepsilon \leq \varepsilon_{max})$;

**If** random$(0,1) > \varepsilon$

  Randomly choose action $a_t$ from action space

**Else**

  $a_t = argmax_{A \in \Delta(A(S))} \sum_{i \in M} Q_i(s, a)$

**End if**

## 5. Case study

### 5.1. *Experimental setup*

The case study of the developed system to optimize an advanced ozonation process for textile color fading with respect to multiple objectives was implemented for the evaluation purpose. Color fading is an essential finishing process for specific textile products such as denim to obtain a worn effect and vintage fashion style [56]. But this effect conventionally was achieved by chemical procedures which have an expensive cost, and highly consume water and power, resulting in heavy negative impacts on the environment. Instead, ozone treatment is an advanced finishing process employing ozone gas to achieve color faded effects on textile products without a water bath, so that save power and water, and causes less environmental issues. The interrelated influences of this process on its process performances have been investigated in our previous works [57–60], and according to the experience data with 129 samples we collected from these experimental studies, four random forests (RF) predictive models were constructed for simulating the 4 process performances of the color fading ozonation process. The present case study will attempt to solve the optimization problems of the color fading ozonation process with regard to the 4 process performance using the DQN based MARL system.

The RF models are inputted by 4 ozonation process parameters (water-content, temperature, *pH* and treating time) to predict four objective color faded performances in terms of color indexes known as *k/s*, $L^*$, $a^*$, and $b^*$ of the treated fabrics with the accuracy of $R^2$=0.996, 0.954, 0.937 and 0.965 respectively. The *k/s* value indicates the color depth, while $L^*$, $a^*$, and $b^*$ are the color indexes from a widely used international standard illustrating the color variation in three dimensions (lightness from 0 to 100, chromatic component from green to red and from blue to yellow from -120 to 120 respectively) [61]. Normally, the color of the final textile product in line with specific *k/s*, $L^*$, $a^*$, and $b^*$ is within the acceptable tolerance of the consumer.

We optimize the color performance in terms of *k/s*, $L^*$, $a^*$, and $b^*$ of the textile in ozonation process by finding a solution including proper parameter variables of water-content, temperature, *pH* and treating time that minimizes the difference between such specific process treated textile product and the targeted sample. Therefore, there are four agents in the stochastic Markov game, and the state space $\varphi$ of it is composed by the solutions containing four parameters (water-content, temperature, *pH* and treating time) in terms of $S_t = [s_t^{v_1}, s_t^{v_2}, s_t^{v_3}, s_t^{v_4}]$. In a time step *t*, given the adjustable units of these parameter variables $u$ = 50, 10, 1, 1 with regard to the constraint ranges of [0, 150],

[0,100], [1, 14] and [1, 60] respectively, as the action of a single variable $v_j$ could be kept (0) or changed up (+) / down (-) in the given range with specific unit $u$, so there are $3^4=81$ actions totally in the action space and the action vector every single agent at time step $t$ is $A_t = [a_t^{v_1}, a_t^{v_2}, a_t^{v_3}, a_t^{v_4}]$, where $a_t^{v_1} \in \{-50, 0, +50\}, v_1 \in [0, 150]; a_t^{v_2} \in \{-10, 0, +10\}, v_2 \in [0, 100]; a_t^{v_3} \in \{-1, 0, +1\}, v_3 \in [1, 14]; a_t^{v_4} \in \{-1, 0, +1\}, v_4 \in [1, 60]$.

The transition probability is 1 for the states in the given range of state space above, but 0 for the states out of it. The reward $r$ of an agent at time step $t$ is expected to be in line with how close the agent gets to its target representing the related objective function. We set up the reward function as illustrated below to induce the agents to approach corresponding optimization objective results:

$$r_t = \sqrt{(f_i(s_t) - p_i)^2} - \sqrt{(f_i(s_{t+1}) - p_i)^2} \quad for \quad i = 1, \ldots m \tag{10}$$

As demonstrated the pseudo-code of DQN based MARL main body in Algorithm 1, the expected color performances of ozonation process treated samples ($p_1, p_2, p_3, p_4$, in terms of $k/s$, $L^*$, $a^*$, and $b^*$) are sampled by experts as 0.81, 15.76, -20.84, and -70.79 respectively to function the system in the present case study. Therefore, there are four agents in this case with respect to their corresponding optimization targets. In addition to the targets, the parameters of DQN agents such as step $F$ for updating Q-networks and replay memory size $D$, as well as the learning rate $\alpha$ and the discount rate $\gamma$ for updating loss function, etc., are listed in Table 1. In particular, the $F$ step for updating DQN here denotes that after 100 steps, the Q-networks would be updated at every 5 steps.

TABLE I. DQN ALGORITHM SETTING IN TEXTILE OZONATION PROCESS CASE STUDY

| $F$ | $D$ | $\alpha$ | $\gamma$ | $\varepsilon_{increment}$ | $a_{max}$ | $E$ | $N$ |
|---|---|---|---|---|---|---|---|
| 5(>100) | 2000 | 0.01 | 0.9 | 0.001 | 0.9 | 1 | 5000 |

The neural networks implemented by TensorFlow [62] are used in our case study to realize Q-networks, and specifically, the networks consist of two layers with 50 and $3^4$ hidden nodes respectively, where the last layer corresponds to the actions. In order to reflect the effectiveness and efficiency of the proposed DQN-based MARL system for multi-objective optimization of the textile manufacturing process in this case study, multi-objective particle swarm optimization (MOPSO). And NSGA- II is considered as the baseline algorithms in this case study due to the popularity of them in the previous applications.

## 5.2. Results and discussion

In the case study, we trained four agents on the basis of the DQN algorithm in a Markov game to optimize an ozone textile process with multiple objectives. As shown in Figure 4 that the increasing $\varepsilon$-greedy policy was used for agents to balance the exploration and exploitation of states. Where the exploration decays in the first 900 steps so that agents initially lack the information and policy explore possible actions, but increasingly follows its policy exploiting the available information by taking action selection mechanism $h$, rather than acting randomly. The effects of it are clearly illustrated on the convergences of DQN agents given in Figure 5 (for the illustration conveniences, 200, 400, and 600 units of loss are additional given to agent 2, agent 3, and agent 4 respectively). It denotes that the deep Q-networks adapts successfully to the stochastic environment that the representation of Q-value in this deep Q-networks for agents is stable and accurate and the agents act deterministically after 900 steps when the $\varepsilon$-greedy increased to the maximum.

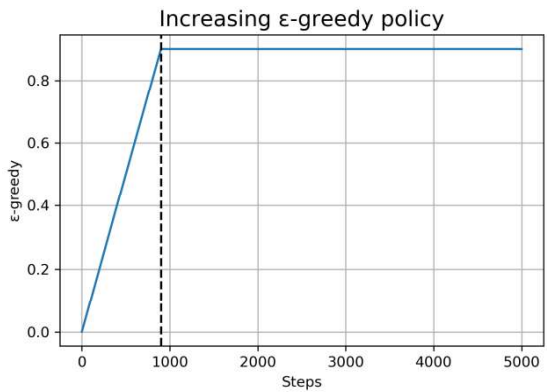

Figure 4. Increasing $\varepsilon$-greedy policy for choosing action

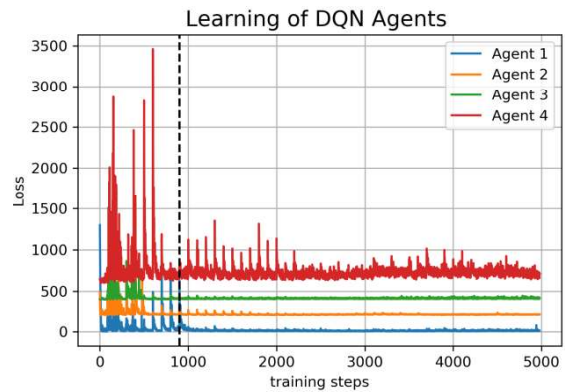

Figure 5. The loss function of DQN for four agents in the Markov game

The agents targeted at optimizing the solution of a textile ozone process to approach the fabric color performance of 0.81, 15.76, -20.84, and -70.79 in regard to $k/s$, $L^*$, $a^*$, and $b^*$. During the DQN agents interacted in the Markov game with 5000 steps, the minimum errors of each agent and their sum in total given by RF models are collected and displayed in Figure 6. The convergence diagrams of all the four agents and their sum in terms of minimum error, verify the effectiveness and efficiency of the designed reward function, and it seems that the solution with lower error can possibly be obtained along with growing time steps.

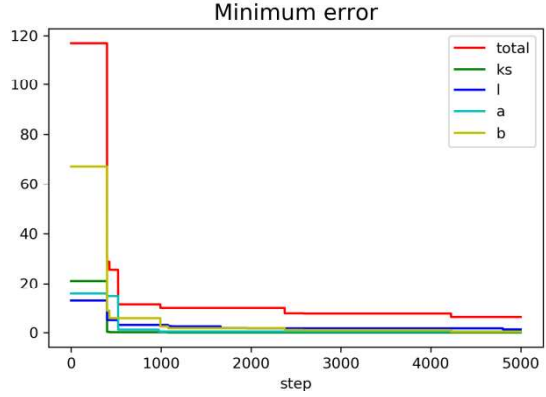

Figure 6. The minimum error of DQN agents tuned and their sum value versus time steps

The comparison of the constructed framework with baseline approaches about optimized results is depicted in Table 2 and Figure 7. The multi-agent reinforcement learning (MARL) system proposed performed dominated the baseline methods of MOPSO and NSGA- II in our case study to optimize the ozonation process solution and achieve the objective color on treated fabrics. The difference from these comparative results could be explained as that the meta-heuristic algorithms of MOPSO and NSGA-2 have been reported that may fail to work with smaller datasets [6] and take an impracticably long time in iteration [63]. But more importantly, though they are effective to deal with low dimension multi-objective optimization problems, the increased stress of selection due to the growing dimension in the problem would decline the effects dramatically when the objectives are more than three.

TABLE II. COMPARISON OF BASELINE ALGORITHMS AND THE PROPOSED FRAMEWORK OF OPTIMIZED RESULT

|  | Targets | MARL | MOPSO | NSGA- II |
|---|---|---|---|---|
| $k/s$ | 0.81 | 1.10 | 0.61 | 0.33 |
| $L^*$ | 15.76 | 14.08 | 20.08 | 25.08 |
| $a^*$ | -20.84 | -25.06 | -37.06 | -43.06 |
| $b^*$ | -70.79 | -70.7 | -78.7 | -85.7 |
| $R^2$ | - | 0.999 | 0.986 | 0.979 |

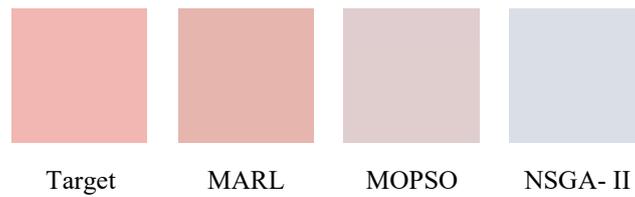

Target　　　MARL　　　MOPSO　　　NSGA- II

Figure 7. Comparison of baseline algorithms and the proposed multi-agent reinforcement learning framework with simulated results

## 6. Conclusions and future work

In this work, we proposed a multi-agent reinforcement learning (MARL) methodology to cope with the increasingly complicated multi-objective optimization problems in the textile manufacturing process. The multi-objective optimization of textile process solutions is modeled as a stochastic Markov game and multiple intelligent agents based on deep Q-networks (DQN) are developed to achieve the correlated equilibrium optimal solutions of the optimizing process. The stochastic Markov game is neither fully cooperative nor fully competitive so that the agents employ a utilitarian selection mechanism that maximizes the sum of all agents' rewards (obeying the increasing ε-greedy policy) in each state to avoid the interruption of multiple equilibria. The case study result reflects that the proposed MARL system is possible to achieve the optimal solutions for the textile ozonation process and it performs better than the traditional approaches.

**Acknowledgments**

This research was supported by the funds from National Key R&D Program of China (Project NO: 2019YFB1706300), and Scientific Research Project of Hubei Provincial Department of Education, China (Project NO: Q20191707).

The first author would like to express his gratitude to China Scholarship Council for supporting this study (CSC, Project NO. 201708420166).